\documentclass[conference]{IEEEtran}
\IEEEoverridecommandlockouts
\usepackage{amsmath,amssymb,amsfonts}
\usepackage{algorithmic}
\usepackage{graphicx}
\usepackage{textcomp}
\usepackage{xcolor}
\usepackage[backend=bibtex,bibstyle=ieee,citestyle=numeric-comp]{biblatex}
\usepackage{caption} 
\def \partha [#1]{\textcolor{red}{Partha: #1}} 
\def\BibTeX{{\rm B\kern-.05em{\sc i\kern-.025em b}\kern-.08em
    T\kern-.1667em\lower.7ex\hbox{E}\kern-.125emX}}
\begin{document}

\title{Attention-Based Methods For Audio Question Answering\\
%{\footnotesize \textsuperscript{*}note: Sub-titles are not captured in Xplore and
}

%\author{\IEEEauthorblockN{Samuel Lipping}
%\IEEEauthorblockA{\textit{Audio Research Group} \\
%\textit{Tampere University}\\
%Tampere, Finland\\
%samuel.lipping@tuni.fi}
%\and

%\IEEEauthorblockN{Parthasaarathy Sudarsanam}
%\IEEEauthorblockA{\textit{Audio Research Group} \\
%\textit{Tampere University}\\
%Tampere, Finland\\
%parthasaarathy.ariyakulamsudarsanam@tuni.fi}
%\and

%\IEEEauthorblockN{Konstantinos Drossos}
%\IEEEauthorblockA{\textit{Audio Research Group} \\
%\textit{Tampere University}\\
%Tampere, Finland\\
%konstantinos.drossos@tuni.fi}
%\and
%\IEEEauthorblockN{Tuomas Virtanen}
%\IEEEauthorblockA{\textit{Audio Research Group} \\
%\textit{Tampere University}\\
%Tampere, Finland\\
%tuomas.virtanen@tuni.fi}
%}

\author{
    \IEEEauthorblockN{ Parthasaarathy Sudarsanam, Tuomas Virtanen}
    \IEEEauthorblockA{\textit{Audio Research Group, Tampere University, Tampere, Finland}}
   % \IEEEauthorblockA{\textit{}}
    \IEEEauthorblockA{\{parthasaarathy.ariyakulamsudarsanam, tuomas.virtanen\}@tuni.fi}
}

\maketitle
\begin{abstract}

Audio question answering (AQA) is the task of producing natural language answers when a system is provided with audio and natural language questions. %Hence, an AQA system must be capable of simultaneously understanding the multimodal inputs as well as learning correlations between them.
In this paper, we propose neural network architectures based on self-attention and cross-attention for the AQA task. The self-attention layers extract powerful audio and textual representations. The cross-attention maps audio features that are relevant to the textual features to produce answers. All our models are trained on the recently proposed Clotho-AQA dataset for both binary yes/no questions and single-word answer questions. Our results clearly show improvement over the reference method reported in the original paper. On the yes/no binary classification task, our proposed model achieves an accuracy of 68.3\% compared to 62.7\% in the reference model. For the single-word answers multiclass classifier, our model produces a top-1 and top-5 accuracy of 57.9\% and 99.8\% compared to 54.2\% and 93.7\% in the reference model respectively. We further discuss some of the challenges in the Clotho-AQA dataset such as the presence of the same answer word in multiple tenses, singular and plural forms, and the presence of specific and generic answers to the same question. We address these issues and present a revised version of the dataset. 

\end{abstract}

\begin{IEEEkeywords}
Audio question answering, attention mechanism, Clotho-AQA
\end{IEEEkeywords}

\section{Introduction}

Question answering (QA) is the task of producing natural language answers when posed with questions in natural language. Often, these questions are accompanied by a natural signal such as an image or audio and the questions posed are about the contents of these signals. If the auxiliary input is an image, the task is referred to as visual question answering (VQA) and if it is an audio signal, it is called audio question answering (AQA). Although the question answering framework is somewhat well-studied for image \cite{YangHGDS15, NEURIPS2019_c74d97b0, Teney_2018_CVPR, fukui-etal-2016-multimodal, https://doi.org/10.48550/arxiv.1606.07356} and textual modalities \cite{devlin-etal-2019-bert, Lan2020ALBERT:, xiong2017dynamic}, audio question answering is comparatively less explored. Audio question answering unlocks new possibilities in areas such as monitoring and surveillance, machine listening, human-technology interaction, acoustical scene understanding, etc. % Developing a system that understands a natural language question and retrieves relevant information from the audio signal is quite challenging. Recent advancements in deep learning for processing multi-modal data make it a suitable candidate to address this task.

%Audio question answering datasets are scarce compared to other modalities like visual question answering ~\cite{Malinowski2014AMA,Zhu2016Visual7WGQ,Agrawal2015VQAVQ,Kafle2017VisualQA,Zhu2016Visual7WGQ,Gao2015AreYT}, textual question answering ~\cite{Rajpurkar2016SQuAD1Q,trischler2017newsqa, yang2018hotpotqa, JoshiTriviaQA2017}. Two popular datasets for AQA are CLEAR ~\cite{abdelnour2018clear}  and DAQA ~\cite{fayek2020temporal}. The CLEAR dataset contains fixed-length audio signals generated from a fixed bank of elementary sounds containing 10 different musical notes. On the other hand, DAQA contains variable-length audio signals generated from 20 different natural sound events such as crowd applauding, dog barking, door slamming, etc. For both these datasets, the questions and answers are generated programmatically such that they test the temporal reasoning capabilities of a system. Although generating questions and answers in a controlled setup reduces errors, they lack the diversity and challenges presented by real data. To overcome this limitation, Clotho-AQA \cite{9909680} dataset was introduced recently for the audio question answering task. This dataset contains variable-length audio signals of everyday environmental sounds such as nature, birds, city, rain, etc. The questions and answers were crowdsourced.

One of the challenging aspects of any multimodal machine learning system is how the information from different modalities is fused to achieve a given task. Traditionally, in question answering systems, the multimodal features are fused using point-wise multiplication \cite{Agrawal2015VQAVQ} or they are concatenated \cite{9909680} to generate a multimodal representation. This may not be an efficient strategy as these features are learned independently without any context from the other modality. Our hypothesis is that using an attention mechanism to learn a multimodal representation helps the model to learn specific features in the audio representation that are closely related to the natural language words in the question and hence improve the performance of the system.

Recently, attention-based architectures \cite{NIPS2017_3f5ee243} have achieved state-of-the-art performances in various tasks ranging from natural language processing \cite{NIPS2017_3f5ee243}, to image classification \cite{dosovitskiy2021an}, sound event detection \cite{miyazaki2020convolution}, sound event localization and detection \cite{https://doi.org/10.48550/arxiv.2107.09388} etc. The ability of transformers to learn powerful and meaningful representations is due to the self-attention and cross-attention layers. Traditional self-attention layers learn bidirectional temporal characteristics of their inputs efficiently. Cross-attention layers are useful to learn or combine features in multimodal translation tasks. For example, in visual question answering, \cite{YangHGDS15, 7780868} used cross-attention layers to improve the ability of the model to find relevant visual cues depending on the question. However, the effect of attention layers for audio question answering task is unexplored.   

In this work, %our contributions are twofold. First, 
we propose neural network architectures based on attention mechanisms and study their effectiveness for the audio question answering task. Our results show improvement over the reference method described for the Clotho-AQA dataset in \cite{9909680}. %Secondly, we also contribute a refined version of the Clotho-AQA dataset which overcomes some of the limitations in the primary version of the dataset. 

The remainder of this paper is organized as follows. In Section \ref{methods}, our proposed method for the AQA task is explained. Then in Section \ref{evaluation}, the dataset, reference methods, and experimental setup are described in detail. Subsequently, in Section \ref{results}, the results of all our experiments are presented. Finally, in Section \ref{conclusion}, the conclusion of this work and possible future works are discussed.

\section{METHODS}\label{methods}

\subsection{Proposed model}
% insert image
\begin{figure}
    \centering
    \includegraphics[width=0.5\textwidth, keepaspectratio]{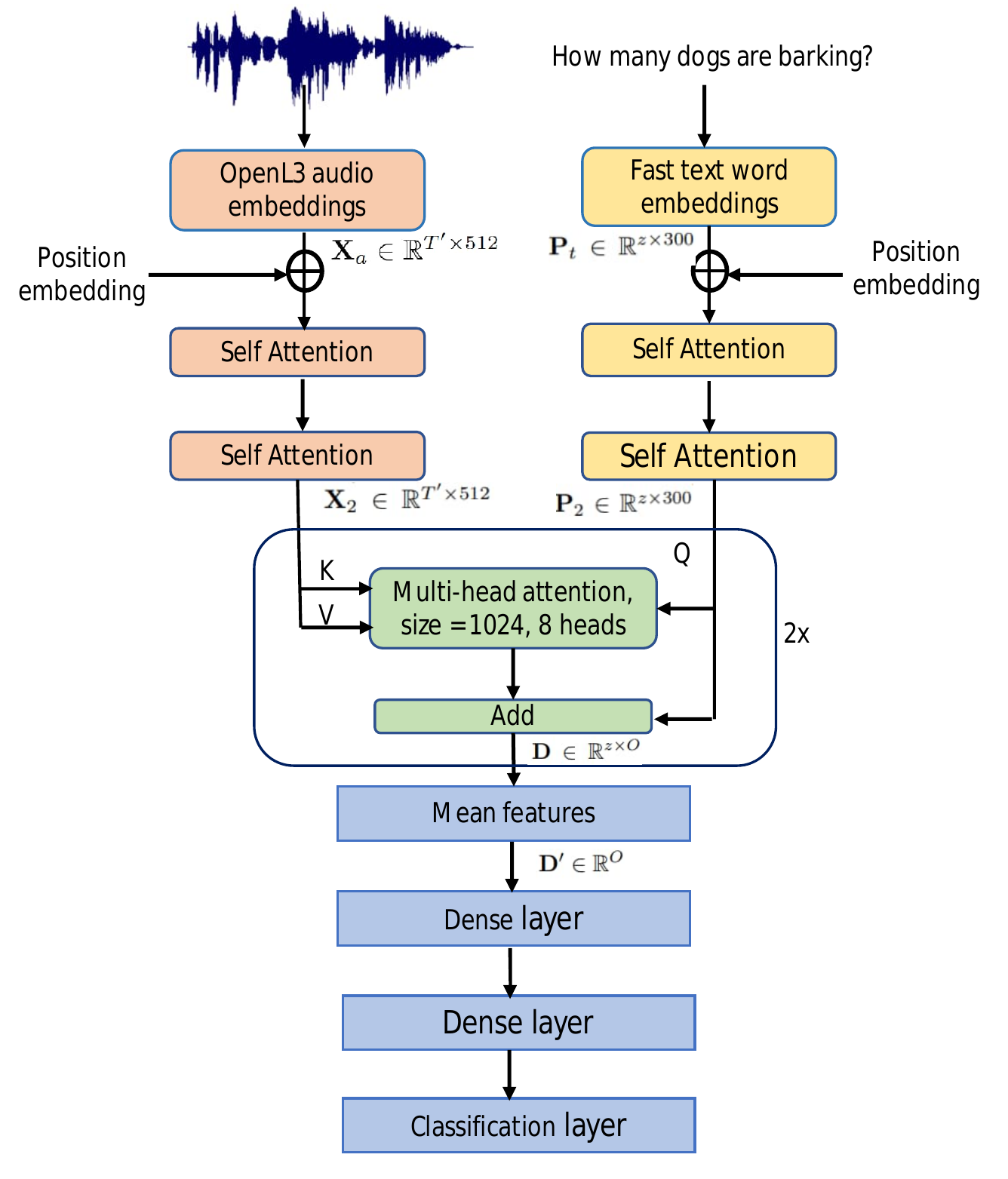}
    \caption[Attention model architecture] {Proposed attention model architecture}
    \label{fig:attention_model}
\end{figure}

An AQA system processes an audio signal and an associated natural language question to produce a natural language answer. The answers can be produced either using a generative model or the answers can be chosen from a list of possible answers using a discriminative classification model. 

In this work, the AQA task is tackled as a classification problem. Our proposed model architecture is shown in Figure \ref{fig:attention_model}. It consists of two input branches, one for processing the audio features and the other for textual features. The audio branch takes in the 
mel-spectrogram of the audio signal with 128 mel bands and $T$ time frames and uses a pre-trained OpenL3 ~\cite{8682475} to extract audio features. The OpenL3 model is based on L\textsuperscript{3}-Net~\cite{DBLP:journals/corr/ArandjelovicZ17} trained on videos from the Audioset~\cite{45857} dataset for audio-visual correspondence task. The output from the pre-trained OpenL3 audio sub network is $\mathbf{X}_{a}\in\mathbb{R}^{T'\times 512}$, where $T'$ is the number of output time frames in the OpenL3 model and 512 is the audio embedding size. 

Similarly, the text branch produces the word vectors of the input textual question % $\mathbf{P}$ 
with $z$ words using the pre-trained Fasttext \cite{mikolov2018advances}. The output representation from the Fasttext model is $\mathbf{P}_{t}\in\mathbb{R}^{z\times 300}$, where 300 is the dimension of the learned word vectors.

These extracted features are passed individually through a series of self-attention (SA) layers for both modalities. In the SA mechanism, each time step in the input feature attends to all other time steps to learn temporal relationships. Since all the time steps are fed simultaneously to the self-attention layer, the order of input features is not known. Hence, we add sinusoidal positional embeddings described in \cite{NIPS2017_3f5ee243} to each of the time steps of the audio and textual features which aids the model in learning the relative positions of these features. 

The SA layers calculate the dot-product attention of the input features with themselves. For any input $\mathbf{H} \in \mathbb{R}^{t\times i} $, where $t$ is the number of time steps and $i$ is the input dimension, the output of the SA layer is calculated as
\begin{equation}
    \text{SA(\textbf{H})} = \text{softmax}\mathbf{(HW_q W_k^TH^T) HW_v}\\
    \label{eq:sa}
\end{equation} 
\noindent
where, $\mathbf{W_q, W_k} \in \mathbb{R}^{i\times k}$ and $\mathbf{W_v} \in \mathbb{R}^{i\times o}$ are learnable query, key and value matrices respectively. Here, $k$ is the key dimension in the attention layer and $o$ is the output dimension. The softmax operation is performed over the rows. Note that in a self-attention layer, the query, key, and value are calculated from the same input.

For the audio features $\mathbf{X}_{a}$ obtained from the OpenL3 model, the output of the $n^{th}$ audio SA layer is $\mathbf{X}_{n}\in\mathbb{R}^{T'\times N}$, where $N$ is the output attention size. Similarly, for the word vectors $\mathbf{P}_{t}$ from the Fasttext model, the output of the $n^{th}$ text SA layer is $\mathbf{P}_{n}\in\mathbb{R}^{z\times M}$, where $M$ is the output attention size. In our experiments, the output attention size of the SA layers in the audio branch is fixed to 512, while that of the text branch is fixed to 300. We used two SA layers for both branches.

To perform a fusion of the audio and textual features, we use multi-head cross-attention (MHCA) layers to learn relevant multimodal features. Attention is applied from the textual features to the audio features to determine which audio features are important to each of the question words. Hence, the output of the final text self-attention layer $\mathbf{P}_{2}\in\mathbb{R}^{z\times 300}$ is used to compute the query and the output of the final audio self-attention layer $\mathbf{X}_{2}\in\mathbb{R}^{T'\times 512}$ is used to calculate the key and value inputs to the MHCA layers. The output of a cross-attention (CA) layer is computed as %If $\mathbf{X}_{2}$ and  are the outputs from the second audio and text self-attention layers respectively, then their cross-attention can be calculated as,

\begin{equation}
     \text{CA}(\mathbf{P_2,X_2}) = \text{softmax}\mathbf{(P_2W_q W_k^TX_2^T) X_2W_v}\\
\end{equation}

%The output of these MHCA layers can be expressed as

%\begin{equation}
%    \mathbf{V}_{m} = \text{MHCA}_{m}({Q}_{m}, K, V)\\
%\end{equation}

%where, $Q_m$ is the query to the $m^{th}$ MHCA layer, $K$, and $V$ are the key and value to the multi-head cross-attention layers respectively. 

\noindent
where, $\mathbf{W_q} \in \mathbb{R}^{300\times 512}$, $\mathbf{W_k} \in \mathbb{R}^{512\times 512}$ and $\mathbf{W_v} \in \mathbb{R}^{512\times O}$ are learnable query, key and value matrices respectively and the softmax operation is performed over the rows. Here, $O$ is the output dimension. For an MHCA layer with M attention heads, the outputs from all the heads are concatenated along the rows and $\mathbf{W_p} \in \mathbb{R}^{MO \times O}$, a learned projection matrix projects it into the desired output dimension. The output of the MHCA layer is given by

 \begin{equation}
    \text{MHCA}\mathbf{(P_2, X_2)} = \underset{m=1...M}{Concat} [\text{CA}_{m}(\mathbf{P_2, X_2})]\mathbf{W_p}\\
\end{equation}

In all our experiments, the output attention size is fixed at 1024 with 8 attention heads. All the hyperparameters were tuned based on the model’s performance on the validation data. The output of the MHCA layer is $\mathbf{D}\in\mathbb{R}^{z\times O}$. To obtain a fixed size representation, the mean is taken over the words axis of the output of the attention layer to produce  $\mathbf{D'}\in\mathbb{R}^{O}$. This is then passed through two dense layers for combining the learned features and then to the classification layer.

We developed two classifiers using this architecture. A binary classifier for questions that have ‘yes' or ‘no' as answers and a multiclass classifier for questions that have other single-word answers in the Clotho-AQA dataset. The classification layer is a logistic regressor with one neuron for the binary classifier. In the case of the multiclass classifier, the final classification layer contains as many neurons as the number of unique answer words in the dataset followed by a softmax activation to predict the probabilities.

\section{EVALUATION} \label{evaluation}

\subsection{Dataset}

We trained and evaluated our models on the recently proposed Clotho-AQA dataset \cite{9909680}. The dataset contains 1991 audio files randomly selected from the Clotho dataset \cite{9052990}. The Clotho dataset is an audio captioning dataset that contains 4981 audio files that are 15-30s in duration. It contains audio files of day-to-day sounds occurring in the environment such as water, nature, birds, noise, rain, city, wind, etc. In the Clotho-AQA dataset, for each of these audio files, there are four ‘yes' or ‘no' questions and two single-word answer questions. For each question, answers were collected from three independent annotators. Hence, each audio file is associated with 18 question-answer pairs. In the Clotho-AQA dataset, there are 828 unique single-word answers excluding ‘yes' and ‘no'. The complete process of data collection, cleaning, and creating the splits is detailed in \cite{9909680}. Henceforth, this dataset is referred to as Clotho-AQA\textunderscore v1 in this paper.

The Clotho-AQA\textunderscore v1 dataset has a few limitations in single-word answers due to crowdsourcing. The dataset contains issues relating to specificity, tense, singular and plural words, etc in the answers. For example, to questions like ‘What is making the chirping sound?', some annotators provided ‘bird' as the answer while some provided ‘seagull' as a more specific answer. Although both these answers can be correct, they are considered different answer classes which creates confusion in the system. Tense issues generally occur when the same question is posed in different tenses for different audio files. For example, for questions like ‘What is the person doing?' and ‘What does the person do?' the answers are ‘running' and ‘run' respectively. An example of singular-plural answers is, for the question ‘Which object is making the metallic sound?', some annotators answered with the singular form ‘key' while some used the plural form ‘keys'. These are considered different answer classes in the Clotho-AQA dataset and thus affect the performance of the classifier.

\begin{figure}
    \centering
    \includegraphics[width=\linewidth]{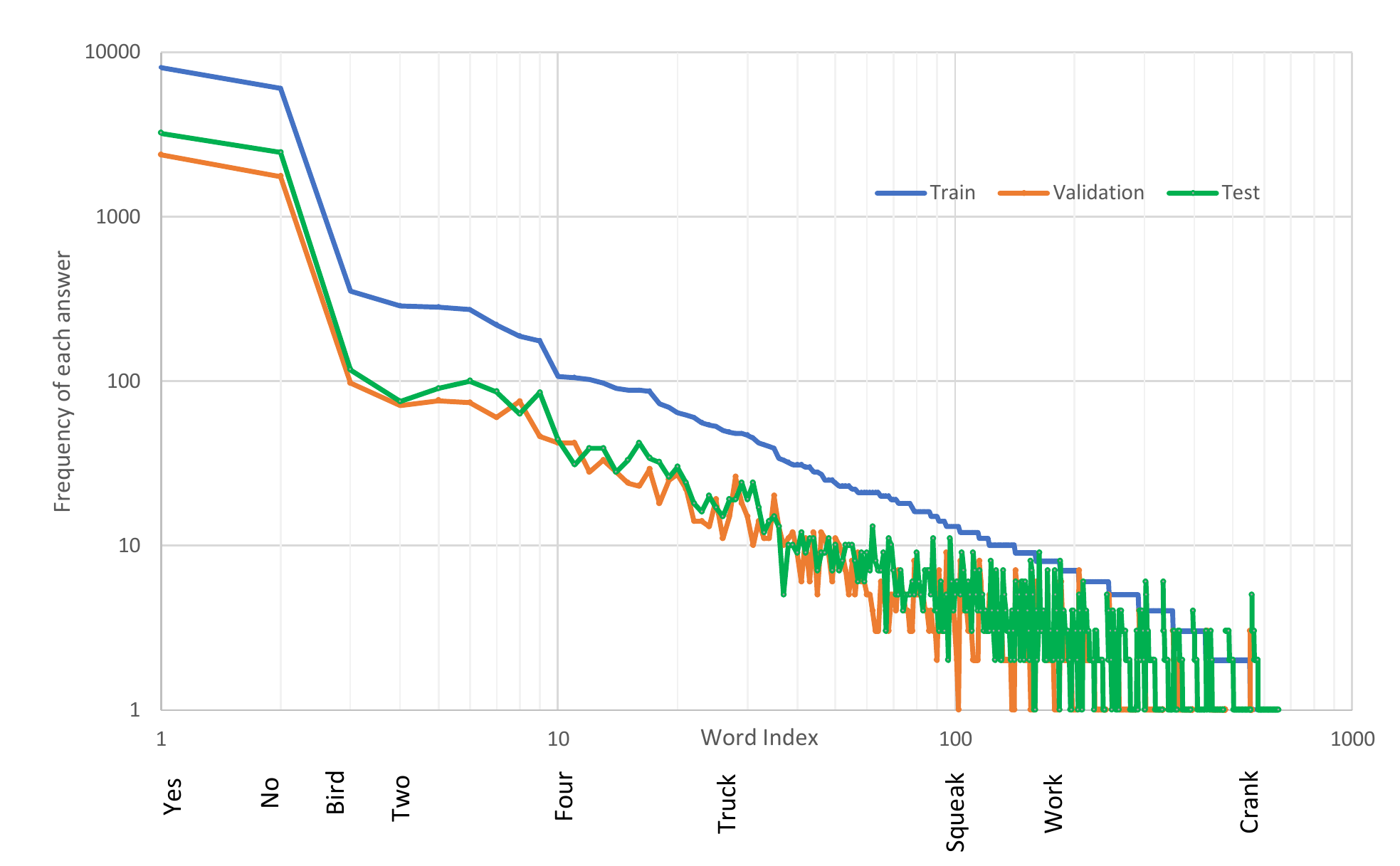}
    \caption[Unique answer words in Clotho-AQA\textunderscore v2] { Count of unique answers in each of the splits in Clotho-AQA\textunderscore v2}
    \label{fig:data_stats_v2}
\end{figure}

Since all the answers in the dataset are single words, the AQA system is trained as a classifier where each unique answer word is denoted by a class index for the ground truth labels. Hence, the system does not learn any language modeling from the answer words. Therefore, we addressed these three issues by replacing specific answer words with their parent classes (for example, ‘seagull' to ‘bird'), all plural words to singular, and all answer tenses to simple present. After this cleaning process, we ended up with 650 unique single-word answers compared to 828 in Clotho-AQA\textunderscore v1. This new version of the dataset is referred to as Clotho-AQA\textunderscore v2 in this paper. The distribution of unique answer words in Clotho-AQA\textunderscore v2 is shown in Figure \ref{fig:data_stats_v2}.

Since each question is answered by three independent annotators, it is also important to analyze if the answers are the same or different from each other. For example in the Clotho-AQA\textunderscore v1 test split for single-word answers, out of 946 unique questions, 203 questions have unanimous answers provided by all the annotators and 381 questions have two out of the three annotators providing the same answer. This means that the maximum possible top-1 accuracy of a system without modeling the characteristics of the annotator will be 61\%. In the Clotho-AQA\textunderscore v2, the maximum achievable accuracy of the multiclass classifier increased to 65\%. 

% insert image

\subsection{Reference methods}
In order to study the effects of cross-attention and self-attention mechanisms on the AQA task, two reference architectures are used. As the first reference model, we used the architecture proposed in \cite{9909680}. This model architecture does not have any attention blocks. Similarly to the proposed model, the mel-spectrogram of the audio input is processed using the pre-trained OpenL3 model and the pre-trained word vectors are obtained using the Fasttext model. Compared to the proposed model, where the pre-trained audio and textual features are passed separately through SA layers, in this model these features are passed separately through a series of Bi-LSTM layers to learn the temporal features and create fixed-size representations. The hidden state of the final time step of the last Bi-LSTM layer is used as the fixed-size representation. These fixed-size representations from the audio and text branches are concatenated to produce the multimodal representation. The multimodal features are then processed by the dense and classification layers similarly to the proposed model. 

The second reference model examines the effect of the cross-attention mechanism on the fusion of multimodal features. MHCA layers are introduced after the Bi-LSTM layers of the first reference system for multimodal representation instead of feature concatenation. The output of the audio Bi-LSTM units is $\mathbf{X}_{audio}\in\mathbb{R}^{T'\times 2h}$, where $h$ is the number of hidden units in the Bi-LSTM layer and $T'$ is the number of output frames from the OpenL3 model. Similarly, the output of the textual branch Bi-LSTM is $\mathbf{X}_{text}\in\mathbb{R}^{z\times 2h'}$, where $h'$ is the number of hidden units in the Bi-LSTM layer and $z$ is the number of words in the natural language question. The cross-attention is calculated on these features as explained in Section \ref{methods}. Similarly to the proposed method, the mean is taken over the words axis to produce a fixed-size representation and it is passed through dense and classification layers for predicting the answer class.

\subsection{Network training}

All the models were trained and evaluated on both the Clotho-AQA\textunderscore v1 and  Clotho-AQA\textunderscore v2 datasets. The data split for the binary classifier and the multiclass classifier is obtained by selecting ‘yes' or ‘no' answers and single-word answers respectively from the Clotho-AQA dataset splits as described in \cite{9909680}. As a result, there are 1174, 344, and 473 audio files for training, validation, and test split respectively each associated with 12 yes/no and  six single-word question-answer pairs.

%The data split for the binary classifier is obtained by selecting the 'yes' or 'no' questions from the respective data splits. As a result, there are 1174, 344, and 473 audio files for training, validation, and test split respectively each associated with 12 'yes' or 'no' question-answer pairs. Similarly, for the multi-class classification, the dataset is obtained by selecting the single-word answers from the respective data splits. This resulted in the same number of audio files as the original splits with each having six question-answer pairs.  

The performance of the binary yes/no classifier is also analyzed on contradicting answers provided by different annotators to the same question similar to the approach proposed in \cite{9909680}. In this regard, three cases are considered. In the first case, all the question-answer pairs are considered valid even if they contain contradicting answers. In the second case, only those question-answer pairs for which all three annotators have responded unanimously are considered valid. In the third case, a majority voting scheme is used, where for each question, the label is chosen as the answer provided by at least two out of the three annotators. These three cases are denoted as ‘Unfiltered data', ‘Unanimous', and ‘Majority votes' respectively. Note that the binary classifier data set is the same in both Clotho-AQA\textunderscore v1 and  Clotho-AQA\textunderscore v2.

All the models were trained for 100 epochs with cross-entropy loss and the model with the best validation score is used for evaluation on the test set.

\section{Results} \label{results}

The results of all the experiments on the Clotho-AQA data set for binary classification of ‘yes' or ‘no' answers are presented in Table \ref{tab:binary_results}. In this table, ‘LSTM' represents the first reference model that uses Bi-LSTM layers for feature extraction and feature concatenation for multimodal fusion. ‘LSTM-CA' represents the model which uses Bi-LSTM layers for feature extraction and cross-attention layers for multimodal fusion. Finally, ‘SA-CA' represents our proposed model which used self-attention layers for feature extraction and cross-attention layers for multimodal fusion. 

Firstly, it is clear that using cross-attention layers significantly improves the performance compared to feature concatenation. This means that the cross-attention layer, helps the model to attend to audio features that are useful to answer the question. Secondly, using self-attention layers with positional embeddings to learn temporal relationships outperform Bi-LSTM layers.

\begin{table}[htbp]
\begin{center}
\begin{tabular}{|c|c|c|c|}
\hline
\textbf{Model} & \textbf{Unfiltered } & \textbf{Unanimous} & \textbf{Majority votes} \\
\hline
\textbf{LSTM} & 62.7 & 73.1 & 63.2\\
\hline
\textbf{LSTM-CA} & 66.2 & 75.4 & 66.3\\
\hline
\textbf{SA-CA} & 68.3 & 77.1  & 68.3 \\
\hline
\textbf{Oracle model} & 86.2 & 100 & 100 \\
\hline
\end{tabular}
\caption[Binary classification results on Clotho-AQA dataset]{Accuracies (\%) of binary ‘yes' or ‘no' classifier on Clotho-AQA data set.}
\label{tab:binary_results}
\end{center}
\end{table}

The results of single-word answer multiclass classifier experiments on Clotho-AQA\textunderscore v1 and Clotho-AQA\textunderscore v2 are summarized in Table \ref{tab:SWC_v1_results} and Table \ref{tab:SWC_v2_results} respectively. Since the number of unique answer classes is large (828 in Clotho-AQA\textunderscore v1 and 650 in Clotho-AQA\textunderscore v2), top-5 and top-10 accuracy scores are also reported. These results indicate that the model is starting to learn relationships between the multimodal data. 

%%%%%%%%%%%%%%%%%%%%%%%%%%%%%%%%%%%%%%%%%%%%%%%%%%%%%%%%%%%%%%%%%
%%%%%%%%%%%%%%%%%%%%%%%%%%%%%%%%%%%%%%%%%%%%%%%%%%%%%%%%%%%%%%%%%

\begin{table}[htbp]
\begin{center}
\begin{tabular}{|c|c|c|c|}
\hline
 \textbf{Model} & \textbf{Top-1 } & \textbf{Top-5} & \textbf{Top-10} \\
\hline
\textbf{LSTM} & 54.2 & 93.7 & 98.0\\
\hline
\textbf{LSTM-CA} & 57.5 & 99.8 & 99.9\\
\hline
\textbf{SA-CA} & 57.9 & 99.8 & 99.9\\
\hline
\textbf{Oracle model} & 61 & 100 & 100 \\
\hline
\end{tabular}
\caption[Single-word answers classification results on Clotho-AQA\textunderscore v1 data set]{Accuracies (\%) of single-word answers classifier on Clotho-AQA\textunderscore v1 data set.}
\label{tab:SWC_v1_results}
\end{center}
\end{table}

%%%%%%%%%%%%%%%%%%%%%%%%%%%%%%%%%%%%%%%%%%%%%%%%%%%%%%%%%%%%%%%%%
%%%%%%%%%%%%%%%%%%%%%%%%%%%%%%%%%%%%%%%%%%%%%%%%%%%%%%%%%%%%%%%%%
\begin{table}[!htbp]
\centering
\begin{tabular}{|c|c|c|c|}
\hline
 \textbf{Model} & \textbf{Top-1 } & \textbf{Top-5} & \textbf{Top-10} \\
\hline
\textbf{LSTM} & 59.8 & 96.6 & 99.3\\
\hline
\textbf{LSTM-CA} & 61.3 & 99.6 & 99.9\\
\hline
\textbf{SA-CA} & 61.9 & 99.8 & 99.9\\
\hline
\textbf{Oracle model} & 65 & 100 & 100 \\
\hline
\end{tabular}
\caption[Single-word answers classification results on Clotho-AQA\textunderscore v2 data set]{Accuracies (\%) of single-word answers classifier on Clotho-AQA\textunderscore v2 data set.}
\label{tab:SWC_v2_results}
\end{table}

It is again evident from the results that the self-attention and cross-attention mechanism significantly improve the evaluation metrics in the case of multi-class classifier as well. It is also noticeable that the model performs better on the  Clotho-AQA\textunderscore v2 dataset after resolving some issues present in the Clotho-AQA\textunderscore v1 dataset.  The proposed multi-class classifier has also reached close to the maximum possible accuracy by an Oracle model  with both Clotho-AQA\textunderscore v1 and Clotho-AQA\textunderscore v2 data sets.

\section{Conclusion} \label{conclusion}

In this paper, we proposed self-attention and cross-attention based architectures for
AQA task. We trained and evaluated 
our models on the recently proposed Clotho-AQA dataset referred 
to as Clotho-AQA\textunderscore v1 in this work. We also discussed some of the 
challenges of this dataset such as the presence of the same answer word in multiple tenses, the presence of specific and generic answers to the same question, and the presence of singular and plural forms of the same word. These challenges were addressed and a revised version of this dataset Clotho-AQA\textunderscore v2 was created. The results of our proposed attention models on both these datasets clearly prove that the cross-attention mechanism helps the model to learn better 
relationships between the input question and the audio compared to the reference methods. It 
is also evident that using self-attention layers with positional embeddings is powerful in learning
useful audio and textual features compared to the Bi-LSTM layers used in the reference methods.
%In the future, we also intend to extend the dataset to reduce data imbalance and minimize
%language biases.
 
\section*{Acknowledgment}

The authors wish to acknowledge CSC-IT Center for Science, Finland, for the computational resources used in this research.

%\section{DATASET} \label{dataset}

%- issues in parent-children labeling \\
%- issues in tenses and singular-plural \\
%- maximum achievable accuracy of V1 of dataset 

%\subsection{Clotho-AQA-v2 dataset}

%- cleaning process \\
%- new dataset statistics \\
%- max achievable accuracy \\
%- issues remaining

%- Two models, cross attention (CA) and self attention (SA) + CA

%\subsection{Model Architecture}\label{AA}

%- describe model architecture with equations and diagrams

%\subsection{Evaluation}

%- training procedure \\
%- describe about binary and multi class classification \\
%- all agree, majority agree, unfiltered

\printbibliography

\end{document}